\documentclass[lettersize,journal]{IEEEtran}
\usepackage{amsmath,amsfonts}
\usepackage{algorithm}
\usepackage{algorithmicx}
\usepackage{array}
\usepackage{textcomp}
\usepackage{stfloats}
\usepackage{url}
\usepackage{verbatim}
\usepackage{graphicx}
\usepackage{cite}
\usepackage[justification=centering]{caption}
\usepackage{algpseudocode}
\usepackage{multirow,booktabs,color,soul}
\usepackage{subfigure}
\usepackage{epsfig}
\usepackage{amssymb}
\usepackage{hyperref}

\begin{document}
	
	\title{DL-DRL: A double-level deep reinforcement learning approach for large-scale task scheduling of multi-UAV}
	
	\author{Xiao Mao, Zhiguang Cao, Mingfeng Fan, Guohua Wu*, and Witold Pedrycz,~\IEEEmembership{Life Fellow,~IEEE}
		\thanks{* Corresponding Author.}
		\thanks{Xiao Mao, Mingfeng Fan and Guohua Wu are with the School of Traffic and Transportation Engineering, Central South University, China (E-mails: maoxiao@csu.edu.cn, mingfan@csu.edu.cn, guohuawu@csu.edu.cn).}
		\thanks{Zhiguang Cao is with the School of Computing and Information Systems, Singapore Management University, Singapore. (E-mail: zhiguangcao@outlook.com).}
		\thanks{Witold Pedrycz is with the Department of Electrical and Computer Engineering, University of Alberta, Edmonton, AB T6G 2V4, Canada, also with the Department of Electrical and Computer Engineering, Faculty of Engineering, King Abdulaziz University, Jeddah 21589, Saudi Arabia, and also with the Systems Research Institute, Polish Academy of Sciences, 01447 Warsaw, Poland. (e-mail: wpedrycz@ualberta.ca).}
	}
	\markboth{}
	{}
	
	\maketitle
	
	\begin{abstract}
		Exploiting unmanned aerial vehicles (UAVs) to execute tasks is gaining growing popularity recently. To solve the underlying task scheduling problem, the deep reinforcement learning (DRL) based methods demonstrate notable advantage over the conventional heuristics as they rely less on hand-engineered rules. However, their decision space will become prohibitively huge as the problem scales up, thus deteriorating the computation efficiency. To alleviate this issue, we propose a double-level deep reinforcement learning (DL-DRL) approach based on a divide and conquer framework (DCF), where we decompose the task scheduling of multi-UAV into task allocation and route planning. Particularly, we design an encoder-decoder structured policy network in our upper-level DRL model to allocate the tasks to different UAVs, and we exploit another attention based policy network in our lower-level DRL model to construct the route for each UAV, with the objective to maximize the number of executed tasks given the maximum flight distance of the UAV. To effectively train the two models, we design an interactive training strategy (ITS), which includes pre-training, intensive training and alternate training. Experimental results show that our DL-DRL performs favorably against the learning-based and conventional baselines including the OR-Tools, in terms of solution quality and computation efficiency. We also verify the generalization performance of our approach by applying it to larger sizes of up to 1000 tasks. Moreover, we also show via an ablation study that our ITS can help achieve a balance between the performance and training efficiency. Our code is publicly available \footnote{https://faculty.csu.edu.cn/guohuawu/zh\_CN/zdylm/193832/list/index.htm}.
	\end{abstract}

	
	\section{Introduction}
	
	Nowadays, unmanned aerial vehicles (UAVs) have been widely applied in the areas of package delivery \cite{liuOptimizationdrivenDynamicVehicle2019}, environment surveillance \cite{machovecDynamicHeuristicsSurveillance2021}, target tracking \cite{altanModelPredictiveControl2020}, and reconnaissance \cite{wangMultiUAVReconnaissanceTask2018} due to their high flexibility, strong mobility, and low power consumption. As the missions to be executed become more and more complex in terms of scale and difficulty, the multi-UAV system has been recognized with salient superiority to the single UAV, where task scheduling plays a vital role to enable the cooperation among the UAVs.  
	
	
	As a variant of the multiple traveling salesman problem (m-TSP) \cite{laporteCuttingPlanesAlgorithm1980}, the multi-UAV task scheduling problem is known to be NP-hard \cite{gerkeyFormalAnalysisTaxonomy2004}, for which it is difficult to find an optimal solution in a short computation time. To solve this problem, both exact and heuristic algorithms have been investigated and explored. Exact algorithms \cite{richardsCoordinationControlMultiple2002, alighanbariCooperativeTaskAssignment2005, forsmoOptimalSearchMission2013} can deliver optimal solutions on small-scale problems, whereas they are relatively incapable on large-scale ones due to the exponentially increasing computation time. Heuristic algorithms, such as genetic algorithm (GA) \cite{edisonIntegratedTaskAssignment2011}, ant colony optimization (ACO) \cite{gaoMultiUAVReconnaissanceTask2021}, particle swarm optimization (PSO) \cite{gengParticleSwarmOptimization2021}, and simulated annealing algorithm (SA) \cite{huoNovelSimulatedAnnealing2020} are desirable to tackle the large-scale problems, which strike a balance between solution quality and computation efficiency. However, the rules in heuristic algorithms always need to be manually designed with domain knowledge and efforts, which may still limit the eventual performance. Additionally, previous studies for the multi-UAV task scheduling usually solve the problem in a whole manner, i.e., directly determining the execution order (or route) of tasks for each UAV, which inevitably runs into the “curse of dimensionality” when solving large-scale problems.
	
	To tackle the large-scale task scheduling for multi-UAV, a number of recent studies decompose the original problem into several subproblems, which are then solved by conventional heuristics~\cite{liuIterativeTwophaseOptimization2021, zitouniDistributedSolutionMultirobot2021}. In this way, the computation complexity could be effectively reduced, thus making the decomposition scheme more favorable for handling large-scale problems. However, the hand-crafted heuristics in those methods may still hold back the performance, as they fail to leverage the underlying pattern among the problem instances to improve the overall performance. On the other hand, with the advances of deep learning (DL) and reinforcement learning (RL), deep reinforcement learning (DRL) has been widely explored in games \cite{mnihHumanlevelControlDeep2015a, zhuOnlineMinimaxNetwork2022}, robotics \cite{zhaoSimtoRealTransferDeep2020}, and natural language processing \cite{luketinaSurveyReinforcementLearning2019}. Recently, DRL is also intensively applied in combinatorial optimization problems (COPs), such as capacitated vehicle routing problem (CVRP) \cite{luLearningbasedIterativeMethod2019}, traveling salesman problem (TSP) \cite{zhengCombiningReinforcementLearning2021}, and orienteering problem (OP) \cite{gamaReinforcementLearningApproach2021a}. In contrast to the conventional heuristics with hand-crafted rules, DRL is able to automatically learn a decision-making policy in a data-driven manner by leveraging the underlying pattern among the COP instances. However, when the problem scales up, the decision space in DRL also expands sharply, which may cause unstable training and undesirable performance \cite{maHierarchicalReinforcementLearning2021}.
	
	To alleviate the issue, in this paper, we exploit a divide and conquer framework (DCF) to decompose the original multi-UAV scheduling problem into a task allocation subproblem and a route planning subproblem. Then, given the decomposed decision space, we propose a double-level deep reinforcement learning approach (DL-DRL) to solve the two subproblems, respectively. In particular, the upper-level DRL model is mainly responsible for the task allocation, in which an encoder based on the self-attention mechanism \cite{vaswaniAttentionAllYou2017} and a UAV selection decoder are designed. The lower-level DRL model shared by all UAVs is mainly responsible for the route planning, where an encoder-decoder structure inspired by the well-known attention model (AM)~\cite{koolATTENTIONLEARNSOLVE2019a} is designed. Unlike the general hierarchical DRL which needs to learn the goal/option determination by trial and error \cite{baconOptionCriticArchitecture2017, nachumDataEfficientHierarchicalReinforcement2018}, the goals of our DL-DRL are determined by the DCF in advance. In this way, the policy learning is reduced to only the actions selection in both levels. Furthermore, we also propose an interactive training strategy (ITS) to train the two DRL models given the potential interplay between them, which includes the pre-training, intensive training, and alternate training. Experimental results justified the effectiveness of our approach in achieving competitive solution quality and computation efficiency, and its desirable generalization to larger problem instances. The main contributions of this paper are summarized as follows. 
	\begin{itemize}
		
		\item A  divide and conquer framework (DCF) is exploited to decompose the multi-UAV task scheduling problem into the task allocation and route planning subproblems. Based on the decomposition, a double-level DRL approach is proposed, in which the encoder-decoder structured policy networks are developed in both the upper-level and lower-level DRL models to solve the two different subproblems, respectively.
		
		\item An interactive training strategy (ITS) is proposed to train the upper-level and lower-level DRL models. This strategy comprises three procedures, including pre-training, intensive training, and alternate training, which could attain desirable balance between training performance and efficiency.
		
		\item Extensive experiments on various scenarios suggest that our DL-DRL approach is able to achieve favorable solution quality and computation efficiency against the learning-based and conventional baselines. The efficacy of our training strategy is also justified through an ablation. Additionally, by applying the model learned for a problem size to solve larger ones (up to 1000 tasks), our DL-DRL approach is verified the desirable generalization performance.
		
	\end{itemize}

	The remainder of this paper is organized as follows. Section II briefly reviews the related works on the multi-UAV task scheduling problem. Section III presents the mathematical formulation of the problem and our divide and conquer framework for problem decomposition. Section IV elaborates our double-level DRL approach and iterative training strategy. Section V conducts extensive experiments and results analysis. Section VI concludes the paper.

	\section{Related Work}
	
	Similar to m-TSP, the multi-UAV task scheduling problem is also a kind of COP with additional constraints like the maximum travelling distance due to the limited battery. Typically, most of the works build a mixed-integer linear program model \cite{richardsCoordinationControlMultiple2002} for this problem and design exact or heuristic algorithms to solve it.
	
	Exact algorithms such as branch and bound \cite{ramirez-atenciaPerformanceEvaluationMultiUAV2015}, branch and price \cite{mitchellMultirobotLongtermPersistent2015}, column generation \cite{mufalliSimultaneousSensorSelection2012}, and dynamic programming \cite{alighanbariCooperativeTaskAssignment2005} have the potential to attain optimal solutions, while they are unsuitable for solving large-scale problems due to the prohibitive computation time for exhaustive search. Thus, heuristic algorithms that could reduce search space with carefully designed heuristic rules are reckoned as desirable alternatives for solving the multi-UAV task scheduling problem. Ye et al. \cite{yeCooperativeTaskAssignment2020} developed an GA method with a multi-type gene chromosome encoding scheme and an adaptive operation for efficient solution searching. Shang et al. \cite{shangGAACOHybridAlgorithm2014} proposed a hybrid algorithm by combining GA and ACO, in which the inferior individuals of the population in GA are replaced with the superior ones in ACO during the population evolution. Chen et al. \cite{chenMultiUAVTaskAssignment2018a} proposed a modified two-part wolf pack search (MTWPS) algorithm based on a designed transposition and extension operation to solve the problem. However, the hand-crafted rules highly rely on the human experience and domain knowledge, hence the solution quality may be undermined.
	
	DRL which takes the advantages of both DL and RL is wildly investigated across various areas like games, robotics, speech recognition, and natural language processing \cite{guptaDeepReinforcementLearning2021} due to its powerful learning capability. As DRL can automatically learn a decision-making policy via the interaction with the environment, it has been recently applied to solve COPs \cite{luo2021real, wangHierarchicalReinforcementLearning2021}, which learns the rules from a large number of problem instances rather than manually designing them. Bello et al. \cite{belloNeuralCombinatorialOptimization2017} presented a DRL model based on pointer networks (PtrNet) \cite{vinyalsPointerNetworks2015} to solve the TSP, which inspired many subsequent works for DRL to solve COPs. To improve the performance of PtrNet-based DRL, Ma et al. \cite{maCombinatorialOptimizationGraph2019a} proposed a graph pointer network that integrates the graph neural network and PtrNet for better feature extraction. In addition, Kool et al. \cite{koolATTENTIONLEARNSOLVE2019a} proposed the AM (Attention Model) method based on the Transformer architecture \cite{vaswaniAttentionAllYou2017} to solve a variety of COPs, which outperforms a wide range of learning-based and conventional baselines. Xu et al. \cite{xuReinforcementLearningMultiple2022} improved the AM to solve VRPs more efficiently by incorporating a multiple relational attention mechanism. These DRL methods exhibit strong performance and reasonable computation efficiency for solving COPs, whereas the training becomes unstable or even out of reach for large-scale problems due to the huge decision space \cite{maHierarchicalReinforcementLearning2021}.
	
	In summary, due to the explosively increasing computational overhead and hand-engineered rules, conventional exact and heuristic algorithms are frustrated in solving the large-scale task scheduling problem of multi-UAV. DRL which automatically learns a decision-making policy through the interaction with the environment (or problem instances) tends to be a desirable alternative. However, the considerable expansion of decision space also hinders the DRL model from training on large-scale problems. To tackle this issue, recent works have concentrated on the problem decomposition. Hu et al. \cite{huReinforcementLearningApproach2020} divided the M-TSP into several small-scale TSP via the proposed DRL policy networks. Liu et al. \cite{liuDeepReinforcementLearning2022} constructed a decomposition framework named EA-DRL to decompose the OP into a knapsack problem and a TSP which are solved by integrating an evolutionary algorithm and DRL, respectively. In this way, the large-scale problem is divided into several small subproblems that DRL can handle efficiently given the relatively small decision space. Inspired by them, we design a DCF that decomposes the large-scale multi-UAV task scheduling problem into two subproblems, i.e., task allocation and UAV route planning. With the DCF, we also propose a double-level DRL approach to solve the two subproblems in the upper level and the lower level, respectively, which has favorable potential for solving the large-scale multi-UAV task scheduling problems.

	\section{Problem Formulation and Framwork Description}
	
	In this section, a mixed-integer linear program model is presented for the multi-UAV task scheduling problem, where the objective is to maximize the total number of tasks completed by UAVs given their maximum flight distance. To solve the large-scale problem, we decompose the original problem into a task allocation subproblem and a UAV route planning subproblem based on the divide and conquer framework (DCF), where the two subproblems are modelled in the upper level and lower level of the DCF, respectively.

	\subsection{Mathematical Formulation}
	
	Let $T=\left\{ 0,1,\cdots ,N \right\}$ be the task set, where 0 and $N$ denote the UAV depot, which a UAV must depart from and finally return to after completing the assigned tasks. Let $U=\left\{ 1,2,\cdots ,V \right\}$ be the UAV set, where $V$ is the number of UAVs. In this paper, all UAVs are considered as homogeneous ones with a same maximum flight distance $D$. Three types of decision variables are defined in the mathematical model of the multi-UAV task scheduling problem, i.e., 1) ${{x}_{ijk}}$ is a binary variable that equals 1 if the UAV $k \in U$ completes the task $j$ right after the task $i$ and 0 otherwise ($i,j \in T$); 2) ${{y}_{ik}}$ is also a binary variable that equals 1 if the UAV $k \in U$ has executed the task $i \in T$ and 0 otherwise; 3) ${{z}_{ik}}$ is a continuous variable indicating the allowed remaining flight distance of the UAV $k\in U$ when visiting task $i\in T$. Accordingly, the mathematical model is formulated as follows,
	
	\begin{small}
		\begin{equation}\label{objectivefunction}
			\begin{array}{cc}
				\text{max}\underset{i=1}{\overset{N-1}{\mathop \sum }}\,\underset{k=1}{\overset{V}{\mathop \sum }}\,{{y}_{ik}}
			\end{array}
		\end{equation}
	\end{small}
	
	Subject to:
	
	\begin{small}
		\begin{equation}\label{constraint1}
			\begin{array}{cc}
				\underset{k=1}{\overset{V}{\mathop \sum }}\,{{y}_{ik}}\le \text{1},i\in \left\{ \text{1},\text{2},\cdots ,N-1 \right\}
			\end{array}
		\end{equation}
		
		\begin{equation}\label{constraint2}
			\begin{array}{cc}
				\underset{j=1}{\overset{N}{\mathop \sum }}\,{{x}_{ijk}}={{y}_{ik}},i\in \left\{ 1,2,\cdots ,N-1 \right\},k\in U
			\end{array}
		\end{equation}
		
		\begin{equation}\label{constraint3}
			\begin{array}{cc}
				\underset{j=1}{\overset{N}{\mathop \sum }}\,{{x}_{ijk}}=\underset{j=0}{\overset{N-1}{\mathop \sum }}\,{{x}_{jik}}\le \text{1},i\in \left\{ \text{0},\text{1},\cdots ,N \right\},k\in U
			\end{array}
		\end{equation}
		
		\begin{equation}\label{constraint4}
			\begin{array}{cc}
				\underset{k=1}{\overset{V}{\mathop \sum }}\,\underset{j=1}{\overset{N-1}{\mathop \sum }}\,{{x}_{0jk}}=\underset{k=1}{\overset{V}{\mathop \sum }}\,\underset{i=1}{\overset{N-1}{\mathop \sum }}\,{{x}_{iNk}}=V
			\end{array}
		\end{equation}
		
		\begin{equation}\label{constraint5}
			\begin{array}{cc}
				\underset{i=0}{\overset{N}{\mathop \sum }}\,\underset{j=0}{\overset{N}{\mathop \sum }}\,{{d}_{ij}}\cdot {{x}_{ijk}} \le D,k\in U
			\end{array}
		\end{equation}
		
		\begin{equation}\label{constraint6}
			\begin{array}{cc}
				{{z}_{ik}}-{{z}_{jk}}+D\cdot {{x}_{ijk}} \le D-{{d}_{ij}},i\ne j\in T,k\in U
			\end{array}
		\end{equation}
		
		\begin{equation}\label{constraint7}
			\begin{array}{cc}
				0 \le {{z}_{ik}} \le D-{{d}_{i0}},i\in \left\{ 1,2,\cdots ,N \right\},k\in U
			\end{array}
		\end{equation}
	\end{small}
	where the objective function (\ref{objectivefunction}) aims to maximize the total number of completed or executed tasks; Constraints (\ref{constraint1}) to (\ref{constraint3}) ensure that each task can be completed once at most; Constraint (\ref{constraint4}) restricts that all the UAVs must depart from and return to the depot; Constraint (\ref{constraint5}) ensures the compliance of the maximum flight distance for each UAV; Subtours are eliminated with constraints (\ref{constraint6}) and (\ref{constraint7}).
	
	\subsection{Divide and Conquer Framework}
	
\begin{figure}[htb]
		\centering{\includegraphics[width=3.6in]{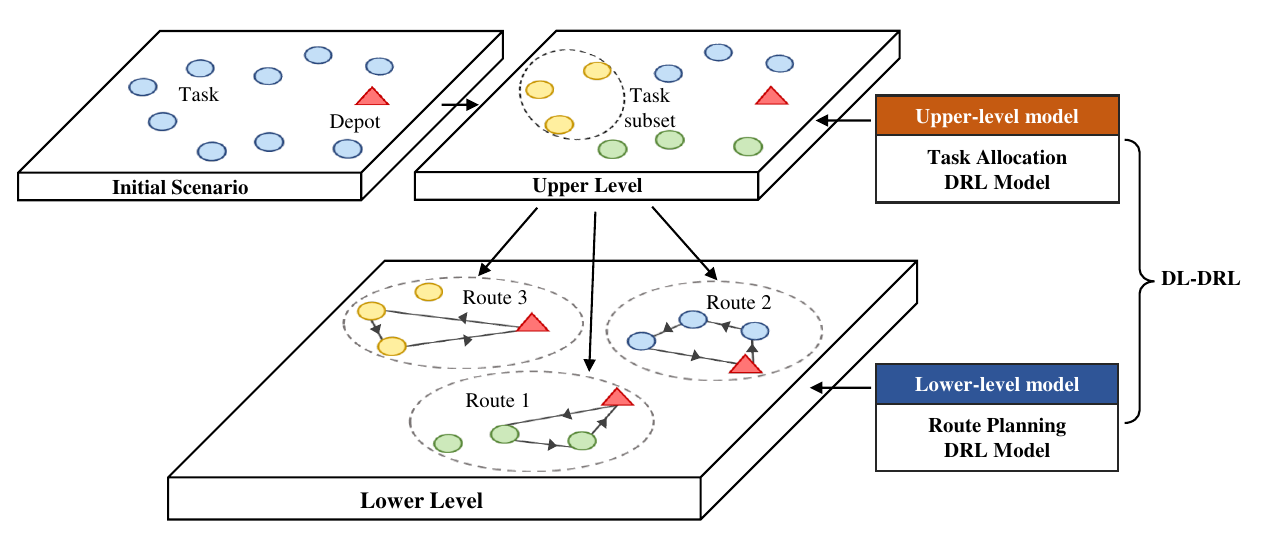}}
		\caption{Divide and conquer framework for task scheduling.}
		\label{Fig. 1}
	\end{figure}

	To reduce the computational complexity of the large-scale multi-UAV task scheduling problem, we decompose it into two subproblems, i.e., task allocation and UAV route planning, based on the divide and conquer framework (DCF). As illustrated in Fig. \ref{Fig. 1}, the DCF comprises an upper level and a lower level. Given a problem instance, at the beginning,  the tasks are allocated to different UAVs in the upper level. Afterwards, each UAV performs the route planning for the allocated tasks in the lower level. However, during the route planning, some tasks are possibly to be excluded out due to the maximum flight distance of the UAV.
	
	Based on this DCF, we propose a DL-DRL approach to solve the multi-UAV task scheduling problem, in which two different DRL models are build in the upper level and lower level, respectively. In the upper level, we design a task allocation DRL model to select an appropriate UAV for each given task. In the lower level, we design a route planning DRL model to construct the route for each UAV, with the objective to maximize the number of tasks to be executed given the maximum flight distance of the UAV. In our DL-DRL, the two models interact with each other, i.e., the upper-level model feeds the allocated tasks as the input to the lower-level model, while the reward of the upper-level model relies on the output of the lower-level model. Through the combination of the models in the two levels, the DL-DRL approach is able to solve the multi-UAV task scheduling problem more efficiently. Moreover, our DL-DRL also has the potential to solve other similar COPs, such as m-TSP \cite{xuTwoPhaseHeuristic2018}, OP \cite{dolinskayaAdaptiveOrienteeringProblem2018}, etc.

	\section{Methodology}
	
	In this section, we introduce our DL-DRL approach for solving the multi-UAV task scheduling problem. For the task allocation subproblem in the upper level, we first cast it as an Markov decision process (MDP) and then design a policy network for the task allocation DRL model. For the UAV route planning subproblem in the lower level, we also design a policy network to construct the route that maximizes the number of tasks to be executed by the UAV given its maximum flight distance. Note that all the UAVs share the same route construction policy in our approach. Afterwards, we propose an interactive training strategy (ITS) that consists of pre-training, intensive training, and alternate training to achieve a balance between the performance and training efficiency.
	
	\subsection{Upper-level DRL Model for Task Allocation}
	
	\subsubsection{Markov Decision Process}
	
	The procedure of the task allocation in the upper level can be deemed as a sequential decision-making process, where a UAV will be assigned for each given task. We cast such a process as an MDP which includes state space $S$, action space $A$, state transition rule $P$, and reward $R$. The detailed definition of our MDP is stated as follows.
	
	\textbf{State.} The state includes the current information of the UAV and task at each step, i.e., ${{s}_{t}}=\left( {{V}_{t}},{{x}_{t}} \right)\in S$, where ${{V}_{t}}$ is the task set that have been allocated to UAVs at time step $t$, and ${{x}_{t}}$ is the task that need to be allocated at time step $t$ and represented using its coordinates. Particularly, ${{V}_{t}}=\left\{ U_{t}^{1},U_{t}^{2},\cdots ,U_{t}^{v} \right\}=$ $\left\{ \{ u_{1}^{1},u_{2}^{1},\cdots ,u_{t}^{1} \},\{ u_{1}^{2},u_{2}^{2},\cdots ,u_{t}^{2} \},\cdots, \left\{ u_{1}^{v},u_{2}^{v},\cdots ,u_{t}^{v} \right\} \right\}$, where $U_{t}^{i}$ and ${u_{t}^{i}}$ are the task set and the corresponding task features (i.e., coordinates of the task) of the UAV $i$ at time step $t$, respectively. 
	
	\textbf{Action.} The action is to allocate the current given task to a UAV, which could be expressed as ${{a}_{t}}=\left( {{u}_{i}},{{x}_{t}} \right)\in A$, i.e., the task ${{x}_{t}}$ will be allocated to the UAV $i$ at time step $t$.
	
	\textbf{State transition rule.} The state transition rule indicates that the state ${{s}_{t}}$ is transited to ${{s}_{t+1}}$ after taking the action ${{a}_{t}}$, i.e., ${{s}_{t+1}}=\left( {{V}_{t+1}},{{x}_{t+1}} \right)=P\left( {{s}_{t}},{{a}_{t}} \right)$. The ${x}_{t+1}$ refers to the next task according to the order of the task sequence in the input instance. The elements in ${{V}_{t}}$ are updated as follows,
	
	\begin{small}
		\begin{equation}\label{state transition rule}
			U_{t+1}^{j}=\left\{ \begin{array}{*{35}{l}}
				\left[ U_{t}^{j},x_{t}^{i} \right],j=i  \\
				\left[ U_{t}^{j},u_{t}^{j} \right],otherwise  \\
			\end{array} \right.
		\end{equation}
	\end{small}
	where $\left[ , \right]$ refers to the concatenation operation, $x_{t}^{i}$ is the task allocated to UAV $i$ at time step $t$, and $u_{t}^{j}$ is the last element in $U_{t}^{j}$ (i.e., the task allocated to UAV $j$ at time step $t$-1). We use this update strategy to keep the same dimension of the allocated task set for each UAV, which may cause repeated tasks for a UAV but would be much convenient for the subsequent batch training.
	
	\textbf{Reward.} Since we aim to maximize the number of tasks that could be executed, we define the reward as the total number of tasks completed by all UAVs, i.e., $R=\underset{i=1}{\overset{v}{\mathop \sum }}\,{{T}_{i}}$, where ${{T}_{i}}$ is the number of tasks completed by UAV $i$, which is obtained based on the route planning DRL model in the lower level.

	\subsubsection{Architecture of the Policy Network}
	
\begin{figure*}[htb]
		\centering{\includegraphics[width=6.8in]{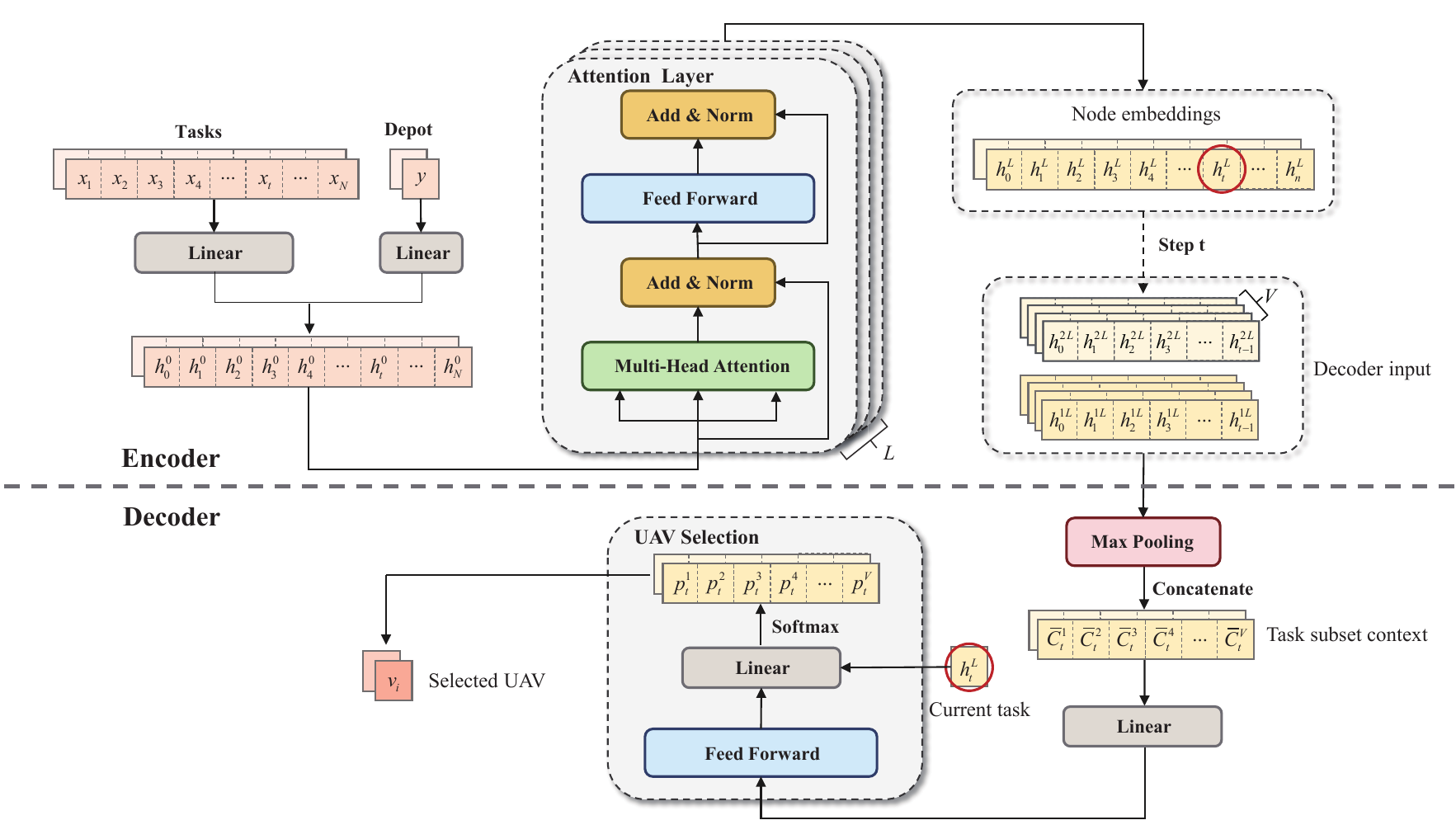}}
		\caption{Architecture of policy network for task allocation in the upper-level DRL model.}
		\label{Fig. 2}
	\end{figure*}

	Compared to the conventional Long Short-Term Memory (LSTM) and Gated Recurrent Unit (GRU) models, the Transformer has stronger capability in handling sequential data, and has been widely employed in domains of natural language processing \cite{tetkoStateoftheartAugmentedNLP2020}, image recognition \cite{dosovitskiyImageWorth16x162020}, recommendation system \cite{sunBERT4RecSequentialRecommendation2019}, and combinatorial optimization \cite{xuReinforcementLearningMultiple2022} as well. Regarding the task allocation, we propose a Transformer-style policy network which is concretized by an encoder and a decoder to allocate a given task to the appropriate UAV, as illustrated in Fig.~\ref{Fig. 2}.
	
	\textbf{Encoder.} The inputs of the encoder are the raw features of tasks and the depot, say the coordinates, and we exploit a linear projection layer and multiple attention layers to learn more informative embeddings. Specifically, the raw inputs are first encoded to ${{d}_{h}}$-dimensional node embeddings ${{h}^{0}}$ via the linear projection with ${{d}_{h}}$=128. Then the node embeddings ${{h}^{0}}$ are promoted to ${{h}^{L}}$ through $L$ attention layers, each of which consists of a multi-head attention (MHA) sublayer and a feed-forward (FF) sublayer.
	
	In the $l\text{-th}$ attention layer, the input ${{h}^{l-1}}=\left\{ h_{1}^{l-1},h_{2}^{l-1},\ldots ,h_{N}^{l-1} \right\}$ is the output of the last attention layer, where $N$ is the number of tasks. To attain the output of the $l\text{-th}$ attention layer ${{h}^{l}}$, the MHA sublayer processes ${{h}^{l\text{-1}}}$ via the multi-head self-attention mechanism first. In specific, for each head, the vectors of $query$, $key$, and $value$ are derived by multiplying ${{h}^{l\text{-1}}}$ and the corresponding trainable parameters. Then, we calculate the attention value ${{Y}_{lm}}$ according to Eq. (11), where $m=\left\{ 1,2,\cdots ,M \right\}$ with $M$=8 and $dim=\frac{128}{M}$. After the calculation of attention values, we concatenate the attention values of all heads and project it to a ${{d}_{h}}$-dimensional vector, and then yield the node embeddings ${{\hat{h}}^{l}}$ through the skip-connection and batch normalization operations. Afterwards, ${{\hat{h}}^{l}}$ is fed to the FF sublayer which is concretized by two linear projections and a ReLu activation function. Here, the skip-connection and batch normalization are also used for the output of the FF sublayer so as to derive the eventual output of the $l\text{-th}$ attention layer. In general, this process could be formulated as follows,
	
	\begin{small}
		\begin{equation}\label{encoder1}
			\begin{array}{cc}
				{{q}_{lm}}=W_{lm}^{Q}{{h}^{l-1}},{{k}_{lm}}=W_{lm}^{K}{{h}^{l-1}},{{v}_{lm}}=W_{lm}^{V}{{h}^{l-1}},
			\end{array}
		\end{equation}
		
		\begin{equation}\label{encoder2}
			\begin{array}{cc}
				{{Y}_{lm}}=\text{softmax}\left( \frac{{{q}^{T}_{lm}}{{k}_{lm}}}{\sqrt{dim}} \right){{v}_{lm}},
			\end{array}
		\end{equation}
		
		\begin{equation}\label{encoder3}
			\begin{array}{cc}
				\text{MHA}\left( {{h}^{l-1}} \right)=\left[ {{Y}_{l1}};{{Y}_{l2}};\ldots ;{{Y}_{lM}} \right]W_{l}^{O},
			\end{array}
		\end{equation}
		
		\begin{equation}\label{encoder4}
			\begin{array}{cc}
				{{\hat{h}}^{l}}=B{{N}^{l}}\left( {{h}^{l-1}}+\text{MHA}\left( {{h}^{l-1}} \right) \right),
			\end{array}
		\end{equation}
		
		\begin{equation}\label{encoder5}
			\begin{array}{cc}
				{{h}^{l}}=B{{N}^{l}}\left( {{{\hat{h}}}^{l}}+\text{FF}\left( {{{\hat{h}}}^{l}} \right) \right),
			\end{array}
		\end{equation}
	\end{small}
	where $W_{lm}^{Q},W_{lm}^{K},W_{lm}^{V}\in{{\mathbb{R}}^{M\times 128\times dim}}$ and $W_{l}^{O}\in {{\mathbb{R}}^{128\times 128}}$ are trainable parameters in the $l\text{-th}$ attention layer.
	
	\textbf{Decoder.} The output of the encoder ${{h}^{L}}$ is used as the input of the decoder. In the decoding process, we choose a task according to the order in the input instance, and then select an appropriate UAV for this task at the current time step and add the task information (i.e., node embedding) to the task subset of the UAV. Specifically, we construct an allocation feature context for UAV $i$, i.e., $\tilde{C}_{t}^{i}=\left[ h_{0}^{iL},h_{1}^{iL},\ldots ,h_{j}^{iL},...,h_{t-1}^{iL} \right]$, where $h_{j}^{iL}$ is the node embedding of the task that is allocated to UAV $i$ at the time step $j$. We aggregate all the allocation feature contexts via a max pooling and concatenate them to obtain a task subset context $\hat{C}_{t}^{TS}$. After that, a linear projection and a 512-dimensional FF layer are used to yield the task subset embeddings $H_{t}^{TS}$. Finally, we concatenate $H_{t}^{TS}$ and node embedding of the current task $h_{t}^{L}$, and then feed it to a linear projection and the softmax function to attain the output probability for task allocation. Based on the probability, we select a UAV for the given task using either greedy decoding or sampling decoding. The formmer always selects the UAV with the highest probability, while the latter selects the UAV by sampling according to its probability. The overall decoding process could be briefly expressed as follows,
	
	\begin{small}
		\begin{equation}\label{decoder1}
			\begin{array}{cc}
				\bar{C}_{t}^{i}=\max \left( \tilde{C}_{t}^{i} \right),
			\end{array}
		\end{equation}
		
		\begin{equation}\label{decoder2}
			\begin{array}{cc}
				\hat{C}_{t}^{TS}=\left[ \bar{C}_{t}^{1},\bar{C}_{t}^{2},\ldots ,\bar{C}_{t}^{V} \right],
			\end{array}
		\end{equation}
		
		\begin{equation}\label{decoder3}
			\begin{array}{cc}
				H_{t}^{TS}=FF({{W}_{1}}\hat{C}_{t}^{TS}+{{b}_{1}}),
			\end{array}
		\end{equation}
		
		\begin{equation}\label{decoder4}
			\begin{array}{cc}
				{{p}_{t}}=\text{softmax(}{{W}_{2}}\left[ H_{t}^{TS},h_{t}^{N} \right]+{{b}_{2}}\text{)},
			\end{array}
		\end{equation}
	\end{small}
	where ${{W}_{1}}$, ${{b}_{1}}$, ${{W}_{\text{2}}}$ and ${{b}_{\text{2}}}$ are trainable parameters.
	
	\subsection{Lower-level DRL Model for Route Planning}
	
	Given the tasks allocated by the upper-level model, we seek to construct a viable route for each UAV to execute those tasks in the lower level, where we design another DRL model to achieve this goal. In light of the nature of route construction, and the notably success achieved by the classic AM \cite{koolATTENTIONLEARNSOLVE2019a} in vehicle routing problems (VRPs), we leverage a similar encoder-decoder structure as the policy network. However, unlike planning a route with the shortest distance in the vanilla AM, we aim to construct a route within the maximum flight distance of the UAV so as to execute as many tasks as possible. In this sense, we set the reward of the lower-level DRL model as the number of executed tasks for a UAV.
	
	Similar to AM, the encoder-decoder architecture is used as the policy network for route construction. In the encoder, coordinates of allocated tasks and depot are first embedded with separate parameters, i.e., utilizing distinct linear projection layers for the initial embedding. After that, attention layers similar to the ones in the upper-level encoder are applied to further learn more informative embeddings for the tasks and depot. To avoid exceeding the maximum flight range of the UAV, we record and update the remaining flying distance during the route planning as follows,
	
	\begin{equation}
		D_{t+1}=D_t-d_{\pi_{t},\pi_{t+1}},
	\end{equation}
	where $D_t$ is the allowed remaining flight distance at time step $t$, $d_{\pi_{t},\pi_{t+1}}$ is the distance between task $\pi_{t}$ and $\pi_{t+1}$, and $\pi_0$ refers to the depot.
	
	In the decoder, we construct the context at time step $t$ with the graph embedding $h_g$, the embedding of the last task $h_{t-1}$ and the remaining flight distance $D_t$. Then the \textit{glimpse} attention \cite{belloNeuralCombinatorialOptimization2017} is applied to the decoder context and compatibilities are calculated. Finally, the probability distribution of executing the task at time step $t+1$ is derived through the softmax operation. Those produces could be briefly expressed as follows,
	
	\begin{equation}
		C_t^l=\left[h_g,h_{t-1},D_t\right],
	\end{equation}
	
	\begin{equation}
		{\widetilde{C}}_t^l=glimpse\left(C_t^l\right),
	\end{equation}
	
	\begin{equation}
		u_t^l=C\cdot tanh(\frac{q_{lt}^Tk_{lt}}{\sqrt{\dim_k}}),
	\end{equation}
	
	\begin{equation}
		p_t^l=softmax(u_t^l),
	\end{equation}
	where $u_t^l$ is the compatibilities calculated by a single-head attention operator;
	$q_{lt}^T={\widetilde{C}}_t^lW_{com}^q$ and $k_{lt}={h^NW}_{com}^k$ are the \textit{query} and \textit{key} vectors for the attention operation, with $h^N$ being the embedding of the tasks, $W_{com}^q$ and $W_{com}^k$ being the trainable parameters; $p_t^l$ is the probability vector of task selection at time step $t$. 
	
	For the convenience of batch training, some tasks might be repeatedly included in the task subset to a UAV, which helps keep the input length the same for the route construction DRL model. Therefore, to shrink the action space of the lower-level DRL model, we impose an initial mask matrix to prohibit the execution of these repetitive dummy tasks. Moreover, to ensure the feasibility of the final solution, during the UAV route construction, tasks that have been executed or cannot be executed within the remaining flight distance will also be masked at each time step.
	
	\subsection{Interactive Training Strategy}
	
	Within the DCF, we designed two different DRL models to solve the task allocation subproblem in the upper-level, and the UAV route planning subproblem in the lower-level, respectively. These two models may affect each other since the input of the lower-level model is determined by the upper-level model, and the reward of the upper-level model is calculated based on the output of the lower-level model. Therefore, we design an interactive training strategy (ITS) that consists of pre-training, intensive training, and alternate training processes to train them effectively and efficiently. The pseudocode of our ITS is presented in Algorithm \ref{algorithm1}. Pertaining to the training algorithm for our two models, we employ the rollout-based REINFORCE which performs well in training Transformer-style DRL models \cite{xuReinforcementLearningMultiple2022, huReinforcementLearningApproach2020, liDeepReinforcementLearning2021c}. Accordingly, the gradient of the loss function in our two models is calculated as follows,
	
	\begin{small}
		\begin{equation}\label{loss}
			\begin{array}{cc}
				\nabla \mathcal{L}\left( \theta |s \right)={{\mathbb{E}}_{{{p}_{\theta }}(\pi |s)}}[(L(\pi )-L({{\pi }^{BL}})){{\nabla }_{\theta }}\log {{p}_{\theta }}(\pi |s)],
			\end{array}
		\end{equation}
	\end{small}
	where $\theta $ is the parameters of the policy network $\pi $, $L(\pi )$ and $L({{\pi }^{BL}})$ are the cost function (i.e., the negative number of executed tasks) of the training model and the baseline model, respectively. We use the greedy decoding strategy in the baseline model. Its parameters will be replaced with that of the current training model if the current training model exhibits significant improvement according to the paired \textit{t}-test. In this way, the loss variance will be potentially decreased during the whole training process. 
	
	\begin{small}
		\begin{algorithm}
			\caption{Interactive training strategy}\label{algorithm1}
			\begin{algorithmic}[1]
				\Require
				the number of pre-training epochs ${{E}_{p}}$;
				the number of model training epochs \textit{E};
				the number of intensive training epochs ${{E}_{t}}$;
				the number of continuous training epochs ${{E}_{c}}$;
				\Ensure
				the training datasets of the pre-training, intensive training and alternate training processes;
				the parameters of networks in the upper-level model ${{\mathrm{M}}_{u}}$ and the lower-level model ${{\mathrm{M}}_{l}}$;
				\For {$i = 1,2,…, {{E}_{p}}$}
				\State $L(\pi_l)\gets -r_l({\mathrm{M}}_{l})$ \Comment{$r_l$ is the reward function}
				\State $L\left(\pi_l^{BL}\right)\gets{-r}_l\left({\mathrm{M}}_{l}^{BL}\right)$
				\State ${{\mathrm{M}}_{l}},{\mathrm{M}}_{l}^{BL}\gets\textbf{REINFORCE}({{\mathrm{M}}_{l}},L(\pi_l),L(\pi_l^{BL}))$
				\EndFor
				\For {$epoch = 1,2,…, \textit{E}$}
				\State $L(\pi_u)\gets -r_u({\mathrm{M}}_{u},{{\mathrm{M}}_{l}})$ \Comment{$r_u$ is the reward function}
				\State $L\left(\pi_u^{BL}\right)\gets{-r}_u\left({\mathrm{M}}_{u}^{BL},{{\mathrm{M}}_{l}}\right)$
				\State ${{\mathrm{M}}_{u},{\mathrm{M}}_{u}^{BL}}\gets\textbf{REINFORCE}({{\mathrm{M}}_{u}},L(\pi_u),L(\pi_u^{BL}))$
				\If {$epoch < {{E}_{t}}$}
				\If {$epoch \% {{E}_{c}} == 0$}
				\State ${\rm TD}_l\gets data\_gen({\mathrm{M}}_{u})$
				\State ${\rm TD}_l\gets shuffle({\rm TD}_l)$
				\State $L(\pi_l)\gets -r_l({\mathrm{M}}_{l})$
				\State $L\left(\pi_l^{BL}\right)\gets{-r}_l\left({\mathrm{M}}_{l}^{BL}\right)$
				\State ${{\mathrm{M}}_{l}},{\mathrm{M}}_{l}^{BL}\gets\textbf{REINFORCE}({{\mathrm{M}}_{l}},L(\pi_l),L(\pi_l^{BL}))$
				\EndIf
				\Else 
				\State ${\rm TD}_l\gets data\_gen({\mathrm{M}}_{u})$
				\State ${\rm TD}_l\gets shuffle({\rm TD}_l)$
				\State $L(\pi_l)\gets -r_l({\mathrm{M}}_{l})$
				\State $L\left(\pi_l^{BL}\right)\gets{-r}_l\left({\mathrm{M}}_{l}^{BL}\right)$
				\State ${{\mathrm{M}}_{l}},{\mathrm{M}}_{l}^{BL}\gets\textbf{REINFORCE}({{\mathrm{M}}_{l}},L(\pi_l),L(\pi_l^{BL}))$
				\EndIf
				\EndFor
			\end{algorithmic}
		\end{algorithm}
	\end{small}
	
	In our ITS, at the beginning of the models training, we construct the pre-training process for the lower-level model (lines 1-5), which is trained for ${{E}_{p}}$ epochs. This process allows the lower-level model to learn an initial route planning policy to support the subsequent training process. After the pre-training process, both the upper-level and lower-level models are trained interactively, with the lower-level model fixed when training the upper-level model and vice versa. During the interactive training, there are two primary processes, 1) intensive training process (lines 7-17), in which the upper-level model is trained for ${{E}_{c}}$ epochs continuously and the lower-level model is trained for the same number of epochs; 2) alternate training process (lines 7-9 and lines 19-23), in which the upper-level model and the lower-level model are trained for one epoch iteratively. In addition, due to the inherent correlation among the multiple instances for the lower-level model, which are generated from the same instance for the upper-level model, we will randomly shuffle the generated instances for the lower-level model to mitigate the correlation.
	
	\section{Experiments and Analysis}
	
	In this section, we conduct extensive experiments to evaluate the performance of our DL-DRL for solving the multi-UAV task scheduling problem, and also compare it with various baselines. The effectiveness of the ITS is also assessed by comparing the convergence of the model with different training strategies. Besides, we further test our DL-DRL on instances of larger-scale to verify its generalization performance.

	\subsection{Experiment Settings}
	
	Following the convention in \cite{koolATTENTIONLEARNSOLVE2019a, liDeepReinforcementLearning2021c, nazariReinforcementLearningSolving2018c}, we randomly generate the locations of tasks and depot in the square $\left[ 0,1 \right]$ following the uniform distribution. We consider two scenarios with 4 and 6 UAVs (termed U4 and U6), respectively, and each scenario includes small-scale instances, medium-scale instances, and large-scale instances according to the number of tasks. Instances with 80 and 100 tasks, 150 and 200 tasks, and 300 and 500 tasks are considered as small-scale, medium-scale, and large-scale, respectively. In this way, we could more comprehensively evaluate the overall performance of the approaches across different sizes of UAVs and tasks. Furthermore, the maximum flight distance of the UAV is set to 2.0 in all instances.

	Regarding our DL-DRL, the number of attention layers in the encoder is set to $L=3$. Both the upper-level and lower-level DRL models are trained for 100 epochs (except  for pre-training) with 1,280,000 randomly generated instances per epoch. The batch size is set to 512 for the model training on small-scale and medium-scale instances, while 256 on the large-scale instances due to the memory constraint. We utilize the Adam optimizer to update the network parameters in the DRL models with an initial learning rate of $10^{-4}$. In the upper-level model, the norm of grads is clipped within 3.0 and the decaying parameter of the learning rate is set to 0.995, while the same two hyperparameters are set to 1.0 in the lower-level model. In the ITS, the lower-level model is pre-trained for 5 epochs (i.e., ${{E}_{p}}=5$), and different parameter settings are used for U4 and U6 scenarios during the interactive training. For the U4 scenario, the number of intensive training epochs ${{E}_{t}}$ is set to 8, and the number of continuous training epochs ${{E}_{c}}$ is set to 4, whereas ${{E}_{t}}$ and ${{E}_{c}}$ are set to 12 and 6 for the U6 scenario, respectively. In this way, the salient fluctuation of the model performance which results from the imbalanced training data of the upper-level and lower-level could be effectively mitigated throughout the training.

	\subsection{Learning Performance}
	
\begin{figure}[htb]
		\vspace{.5pc}
		\begin{center}
			\subfigure[U4-80]{\psfig{file=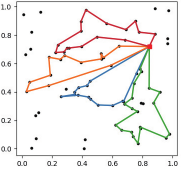,width=1in}}\mbox{ }
			\subfigure[U4-100]{\psfig{file=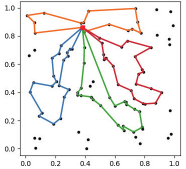,width=1in}}\mbox{ }
			\subfigure[U4-150]{\psfig{file=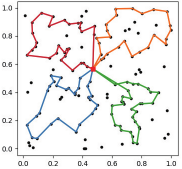,width=1in}}\mbox{ }
			\subfigure[U4-200]{\psfig{file=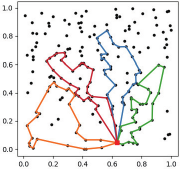,width=1in}}\mbox{ }
			\subfigure[U4-300]{\psfig{file=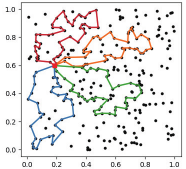,width=1in}}\mbox{ }
			\subfigure[U4-500]{\psfig{file=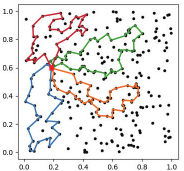,width=1in}}\mbox{ }
			\subfigure[U6-80]{\psfig{file=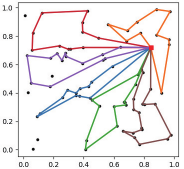,width=1in}}\mbox{ }
			\subfigure[U6-100]{\psfig{file=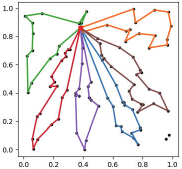,width=1in}}\mbox{ }
			\subfigure[U6-150]{\psfig{file=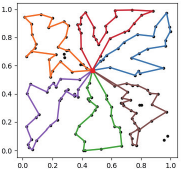,width=1in}}\mbox{ }
			\subfigure[U6-200]{\psfig{file=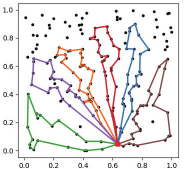,width=1in}}\mbox{ }
			\subfigure[U6-300]{\psfig{file=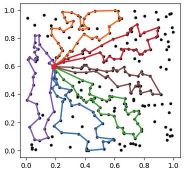,width=1in}}\mbox{ }
			\subfigure[U6-500]{\psfig{file=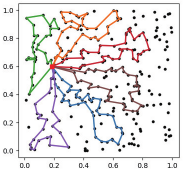,width=1in}}\mbox{ }
			
		\end{center}
		\caption{Solution visualization of trained models in corresponding situations}\label{Fig. 3}
	\end{figure}

	Following the above experiment settings, our DL-DRL model is trained on U4 and U6 scenarios with different scales. For the training on the small-scale and medium-scale instances, a single RTX 3090 GPU with 24GB memory is used, whereas two GPUs are used for the training on the large-scale instances. The training time of the upper-level model, lower-level model, and data generation for the lower-level model for each epoch is displayed in Table \ref{Table 1}. In particular, “U4-80” means the model is trained on the U4 scenario with 80 tasks. Then we visualize the results of trained models on a number of exemplary instances in Fig. \ref{Fig. 3}, where all the instances are randomly generated with seed 1234. Obviously, we can see that our trained DL-DRL models could deliver reasonably good solutions in various situations.
	
	\begin{table}[!ht]
		\vspace{-.5pc}
		\renewcommand\arraystretch{1.2}
		\centering
		\caption{Time Consuming of Model Training Per Epoch (minutes)}
		\begin{tabular}{cccc}
			\toprule
			Situation & Upper-level & Lower-level & Data generation \\ \midrule
			U4-80 & 20.55 & 6.12  & 13.02  \\
			U4-100 & 24.58  & 6.58  & 14.06  \\
			U4-150 & 32.15  & 8.78  & 19.30  \\ 
			U4-200 & 55.98  & 11.15  & 22.77  \\
			U4-300 & 123.70 & 23.32 & 45.50 \\ 
			U4-500 & 206.37 & 31.28 & 84.95 \\ \midrule
			U6-80 & 25.92 & 6.03 & 17.19 \\ 
			U6-100 & 30.65 & 6.63 & 18.58 \\ 
			U6-150 & 49.25 & 8.55  & 23.35  \\ 
			U6-200 & 72.10  & 11.32  & 29.15  \\
			U6-300 & 155.55 & 22.92 & 58.49 \\ 
			U6-500 & 262.43 & 31.27 & 116.02 \\ \bottomrule
		\end{tabular}
		\label{Table 1}
	\end{table}
	
\begin{figure*}[htb]
		\vspace*{-1pc}
		\centering{\includegraphics[width=6.8in]{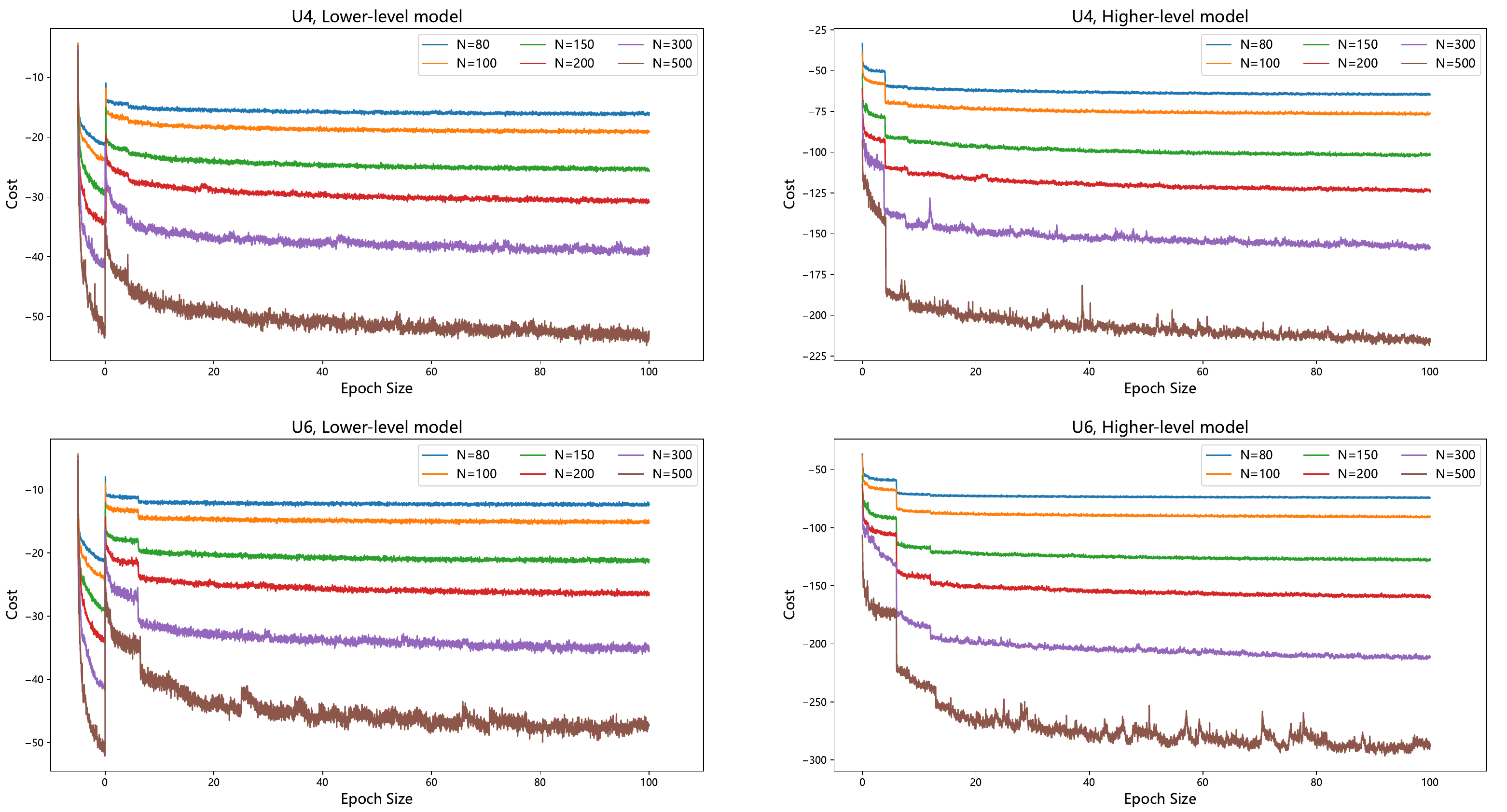}}
		\caption{Learning curves of DL-DRL}
		\label{Fig. 4}
		\vspace*{-.5pc}
	\end{figure*}

	Meanwhile, we plot the learning curves of the upper-level and lower-level models during the training process in Fig. \ref{Fig. 4}. The vertical axis refers to the cost value (i.e., negative of the total number of executed tasks), and the horizontal axis refers to the number of epoch. From Fig. \ref{Fig. 4}, we can observe that, 1) For the lower-level model, it converges fast during the pre-training process, while the cost value dramatically increases at the beginning of the interactive training. This phenomenon might be caused by the difference in the training data distribution between pre-training and interactive training, i.e., the former is a uniform distribution, and the latter is unknown prior and produced by the upper-level model; 2) For the upper-level model, it exhibits several rapid dropping of cost value in the early training period. As the lower-level model performs better and better on the training data produced by the upper-level model during the intensive training process, the upper-level model also quickly delivers good performance. In addition, the learning curves demonstrate more obvious fluctuations on large-scale instances due to the expansion of the state and action space. Our DL-DRL models finally converge stably, suggesting that they have learned valid policies.
	
	\subsection{Comparison Analysis}
	
	We compare the proposed DL-DRL approach with the exact solver, conventional heuristics, and learning-based baselines, which are briefly introduced as follows.
	
	1) Gurobi\footnote{https://www.gurobi.com/}: A commercial exact solver for (mixed) integer programmings; 
	
	2) OR-Tools\footnote{https://developers.google.cn/optimization/}: A widely used strong heuristic solver developed by Google;
	
	3) K-means+VND: Based on the same DCF, it uses K-means and variable neighborhood descent (VND)\footnote{https://github.com/nchlpz/VND \_TOP} to handle task allocation and UAV route planning, respectively;
	
	4) K-means+AM: Based on the same DCF, it uses K-means and AM \cite{koolATTENTIONLEARNSOLVE2019a} to handle task allocation and UAV route planning, respectively;
	
	5) DL-DRL-I: An approach with the same network architecture as our DL-DRL, while it trains the two DRL models independently rather than using the proposed ITS.
	
	As an exact solver, Gurobi is computationally time-consuming for solving multi-UAV task scheduling problems even on small-scale instances. Therefore, for each problem size, we evaluate Gurobi with 30 testing instances and limit the maximum solving time to 1800s, whereas other baselines are evaluated with 500 testing instances without any time limitation. All of the baselines are implemented in Python. The Intel Xeon Gold 5218R with 20 cores and RTX 3090 GPU are utilized for conventional and learning-based approaches, respectively.
	Regarding the upper-level model of DL-DRL and DL-DRL-I, both greedy and sampling strategies are employed in the decoder during the testing. According to Eq. (\ref{decoder4}), the greedy strategy always selects the UAV with the highest probability, whereas the sampling strategy engenders 128 solutions by sampling and reports the best. The experimental results on the U4 and U6 scenarios are presented in Table II and Table III, respectively. These tables gather the average objective value (Obj.), gap, and the average computation time (or runtime) of all methods. Here, the Obj. is the total number of completed tasks on average, and the gap is defined as the normalized distance between an average objective value $Obj$ and the best one $Ob{{j}_{best}}$ found among all methods, which is expressed as follows,
	
	\begin{small}
		\begin{equation}\label{gap}
			Gap=\frac{Ob{{j}_{best}}-Obj}{Ob{{j}_{best}}}\times 100\%.
		\end{equation}
	\end{small}
	
	\begin{table*}[htbp]
		\renewcommand\arraystretch{1.2}
		\footnotesize
		\centering
		\caption{Experiment Results on the U4 Scenario}
		\resizebox{\textwidth}{!}{
			\begin{tabular}{p{4em}<{\centering}p{3em}<{\centering}p{4.25em}<{\centering}p{4.25em}<{\centering}p{5em}<{\centering}|p{3em}<{\centering}p{4.25em}<{\centering}p{4.25em}<{\centering}p{5em}<{\centering}p{5em}<{\centering}p{4.25em}<{\centering}p{5em}<{\centering}}
				\toprule
				Number of tasks & \multicolumn{1}{p{3em}}{} & \multicolumn{1}{p{4.25em}}{Gurobi (1800s) \centering} & \multicolumn{1}{p{4.25em}}{DL-DRL (Greedy) \centering} & \multicolumn{1}{p{5em}|}{DL-DRL (Sampling) \centering} & \multicolumn{1}{p{3em}}{OR-tools \centering} & \multicolumn{1}{p{4.25em}}{K-means +VND \centering} & \multicolumn{1}{p{4.25em}}{K-means +AM \centering} & \multicolumn{1}{p{5em}}{DL-DRL-I (Greedy) \centering} & \multicolumn{1}{p{5em}}{DL-DRL-I  (Sampling) \centering} & \multicolumn{1}{p{4.25em}}{DL-DRL (Greedy) \centering} & \multicolumn{1}{p{5em}}{DL-DRL (Sampling) \centering} \\
				\midrule
				\multirow{3}[2]{*}{80} & Obj.  & 56.77 & 63.67 & \textbf{64.30} & \textbf{65.44} & 56.45 & 59.68 & 58.78 & 61.32 & 64.67 & 65.28 \\
				& Gap   & 11.72 & 0.98  & \textbf{0.00} & \textbf{0.00} & 13.74 & 8.79  & 10.18 & 6.29  & 1.17  & 0.24 \\
				& Time(s) & 1800.00 & 0.03  & 0.11  & 0.38  & 39.20 & 0.09  & 0.00  & 0.12  & 0.00  & 0.11 \\
				\midrule
				\multirow{3}[2]{*}{100} & Obj.  & 65.37 & 76.20 & \textbf{76.87} & 76.83 & 62.65 & 69.35 & 67.61 & 71.14 & 76.67 & \textbf{77.39} \\
				& Gap   & 14.96 & 0.87  & \textbf{0.00} & 0.72  & 19.05 & 10.38 & 12.63 & 8.07  & 1.01  & \textbf{0.00} \\
				& Time(s) & 1800.00 & 0.03  & 0.13  & 0.60  & 53.31 & 0.10  & 0.00  & 0.15  & 0.00  & 0.14 \\
				\midrule
				\multirow{3}[2]{*}{150} & Obj.  & 75.10 & 102.57 & \textbf{103.63} & 101.34 & 74.11 & 87.14 & 90.28 & 96.41 & 102.25 & \textbf{103.42} \\
				& Gap   & 27.53 & 1.03  & \textbf{0.00} & 2.01  & 28.34 & 15.74 & 12.71 & 6.77  & 1.12  & \textbf{0.00} \\
				& Time(s) & 1800.00 & 0.03  & 0.21  & 1.25  & 67.79 & 0.11  & 0.00  & 0.21  & 0.00  & 0.21 \\
				\midrule
				\multirow{3}[2]{*}{200} & Obj.  & 74.37 & 74.37 & \textbf{125.07} & 121.21 & 81.47 & 100.85 & 106.68 & 115.34 & 123.76 & \textbf{124.79} \\
				& Gap   & 40.54 & 40.54 & \textbf{0.00} & 2.87  & 34.72 & 19.19 & 14.51 & 7.57  & 0.83  & \textbf{0.00} \\
				& Time(s) & 1800.00 & 1800.00 & 0.29  & 2.00  & 96.51 & 0.12  & 0.00  & 0.30  & 0.00  & 0.27 \\
				\midrule
				\multirow{3}[2]{*}{300} & Obj.  & 61.77 & 61.77 & \textbf{159.23} & 155.60 & 94.86 & 124.77 & 134.42 & 145.41 & 159.25 & \textbf{160.14} \\
				& Gap   & 61.21 & 61.21 & \textbf{0.00} & 2.84  & 40.77 & 22.09 & 16.06 & 9.20  & 0.56  & \textbf{0.00} \\
				& Time(s) & 1800.00 & 1800.00 & 0.44  & 4.41  & 164.32 & 0.14  & 0.00  & 0.42  & 0.00  & 0.41 \\
				\midrule
				\multirow{3}[2]{*}{500} & Obj.  & 69.33 & 69.33 & \textbf{222.43} & 209.00 & 113.88 & 154.71 & 170.20 & 188.78 & 216.66 & \textbf{220.44} \\
				& Gap   & 68.83 & 68.83 & \textbf{0.00} & 5.19  & 48.34 & 29.82 & 22.79 & 14.36 & 1.72  & \textbf{0.00} \\
				& Time(s) & 1800.00 & 1800.00 & 0.89  & 11.71 & 316.45 & 0.15  & 0.01  & 0.84  & 0.01  & 0.85 \\
				\bottomrule
			\end{tabular}%
		}
		\label{Table 2}%
	\end{table*}%

	\begin{table*}[htbp]
		\renewcommand\arraystretch{1.2}
		\footnotesize
		\centering
		\caption{Experiment Results on the U6 Scenario}
		\resizebox{\textwidth}{!}{
			\begin{tabular}{p{4em}<{\centering}p{3em}<{\centering}p{4.25em}<{\centering}p{4.25em}<{\centering}p{5em}<{\centering}|p{3em}<{\centering}p{4.25em}<{\centering}p{4.25em}<{\centering}p{5em}<{\centering}p{5em}<{\centering}p{4.25em}<{\centering}p{5em}<{\centering}}
				\toprule
				Number of tasks & \multicolumn{1}{p{3em}}{} & \multicolumn{1}{p{4.25em}}{Gurobi (1800s) \centering} & \multicolumn{1}{p{4.25em}}{DL-DRL (Greedy) \centering} & \multicolumn{1}{p{5em}|}{DL-DRL (Sampling) \centering} & \multicolumn{1}{p{3em}}{OR-tools \centering} & \multicolumn{1}{p{4.25em}}{K-means +VND \centering} & \multicolumn{1}{p{4.25em}}{K-means +AM \centering} & \multicolumn{1}{p{5em}}{DL-DRL-I (Greedy) \centering} & \multicolumn{1}{p{5em}}{DL-DRL-I  (Sampling) \centering} & \multicolumn{1}{p{4.25em}}{DL-DRL (Greedy) \centering} & \multicolumn{1}{p{5em}}{DL-DRL (Sampling) \centering} \\
				\midrule
				\multirow{3}[2]{*}{80} & Obj.  & 64.37 & 73.10 & \textbf{73.53} & \textbf{75.42} & 68.59 & 68.31 & 68.91 & 71.01 & 74.00 & 74.32 \\
				& Gap   & 12.47 & 0.59  & \textbf{0.00} & \textbf{0.00} & 9.05  & 9.43  & 8.63  & 5.85  & 1.88  & 1.46 \\
				& Time(s) & 1800.00 & 0.03  & 0.15  & 0.50  & 20.65 & 0.14  & 0.00  & 0.13  & 0.00  & 0.14 \\
				\midrule
				\multirow{3}[2]{*}{100} & Obj.  & 73.23 & 90.27 & \textbf{91.23} & \textbf{92.42} & 81.38 & 82.12 & 82.60 & 85.43 & 90.72 & 91.53 \\
				& Gap   & 19.73 & 1.06  & \textbf{0.00} & \textbf{0.00} & 11.95 & 11.14 & 10.63 & 7.56  & 1.84  & 0.96 \\
				& Time(s) & 1800.00 & 0.03  & 0.19  & 0.85  & 36.32 & 0.16  & 0.00  & 0.19  & 0.00  & 0.17 \\
				\midrule
				\multirow{3}[2]{*}{150} & Obj.  & 79.63 & 128.23 & \textbf{129.60} & \textbf{130.09} & 104.49 & 110.14 & 110.53 & 116.29 & 128.18 & 129.59 \\
				& Gap   & 38.56 & 1.06  & \textbf{0.00} & \textbf{0.00} & 19.67 & 15.33 & 15.04 & 10.60 & 1.46  & 0.38 \\
				& Time(s) & 1800.00 & 0.03  & 0.27  & 2.21  & 88.55 & 0.19  & 0.00  & 0.26  & 0.01  & 0.24 \\
				\midrule
				\multirow{3}[2]{*}{200} & Obj.  & 73.27 & 157.37 & \textbf{159.97} & 160.79 & 117.54 & 136.05 & 137.09 & 144.78 & 158.96 & \textbf{160.83} \\
				& Gap   & 54.20 & 1.63  & \textbf{0.00} & 0.03  & 26.92 & 15.41 & 14.76 & 9.98  & 1.16  & \textbf{0.00} \\
				& Time(s) & 1800.00 & 0.04  & 0.37  & 3.73  & 151.36 & 0.23  & 0.00  & 0.35  & 0.01  & 0.35 \\
				\midrule
				\multirow{3}[2]{*}{300} & Obj.  & 79.80 & 211.93 & \textbf{214.27} & 212.46 & 135.89 & 165.10 & 178.61 & 191.23 & 213.27 & \textbf{215.47} \\
				& Gap   & 62.76 & 1.09  & \textbf{0.00} & 1.40  & 36.94 & 23.38 & 17.11 & 11.25 & 1.02  & \textbf{0.00} \\
				& Time(s) & 1800.00 & 0.05  & 0.58  & 8.00  & 221.86 & 0.28  & 0.01  & 0.57  & 0.01  & 0.53 \\
				\midrule
				\multirow{3}[2]{*}{500} & Obj.  & 61.00 & 295.40 & \textbf{299.73} & 292.10 & 162.25 & 193.21 & 201.76 & 226.21 & 293.18 & \textbf{297.59} \\
				& Gap   & 79.65 & 1.45  & \textbf{0.00} & 1.84  & 45.48 & 35.07 & 32.20 & 23.99 & 1.48  & \textbf{0.00} \\
				& Time(s) & 1800.00 & 0.06  & 1.18  & 20.84 & 382.92 & 0.32  & 0.01  & 1.07  & 0.01  & 1.05 \\
				\bottomrule
			\end{tabular}%
		}
		\label{Table 3}%
	\end{table*}%
	
	Regarding the testing on the 30 instances and 500 instances, the results are displayed in the left half and right half of the two tables, respectively. We can see from Tables \ref{Table 2} and \ref{Table 3} that Gurobi is unable to attain the optimal solution within the given time limit, even on instances with only 80 tasks. At the same time, our DL-DRL always achieves better solutions than Gurobi no matter which decoding strategy is used.
	
	Pertaining to the U4 scenario in Table \ref{Table 2}, it is obvious that our DL-DRL (Greedy) outperforms K-means+VND and K-means+AM in terms of Obj. and computation time, with this superiority becoming more significant as the task scales up. Regarding the DL-DRL and DL-DRL-I, which have the same network architectures, the sampling strategy always yields better solutions than the greedy one, with only slightly longer computation time. Our DL-DRL (Greedy) outperforms DL-DRL-I (Greedy) and DL-DRL-I (Sampling), which implies the effectiveness of the ITS in improving the model performance. In comparison with the strong heuristic solver (i.e., OR-Tools), our DL-DRL (Sampling) is slightly inferior in terms of Obj. on the instances with the fewest tasks. However, as the task scales up, DL-DRL (Sampling) outperforms OR-Tools in terms of both Obj. and computation time, which justified the strong capability of our approach in tackling large-scale task scheduling problems of multi-UAV.
	
	Pertaining to the U6 scenario in Table \ref{Table 3}, a similar pattern could be observed that our DL-DRL (Greedy and Sampling) outperforms K-means+VND, K-means+AM, DL-DRL-I (Greedy), and DL-DRL-I (Sampling). Meanwhile, our DL-DRL (Sampling) also exhibits a competitive performance against OR-Tools. On instances with no more than 150 tasks, DL-DRL (Sampling) is slightly inferior to OR-Tools in terms of Obj., but its computation time is much shorter. When the number of tasks increases, our DL-DRL (Sampling) outstrips OR-Tools in terms of both Obj. and computation time. Combining the results from both tables, DL-DRL demonstrates a better overall performance than Gurobi, K-means+VND, K-means+AM, and DL-DRL-I. It also outperforms OR-Tools on large-scale instances and performs competitively against OR-Tools with shorter computation time on small-scale instances.

	\subsection{Ablation Study on ITS}
	
\begin{figure}[htb]
		\centering{\includegraphics[width=3.5in]{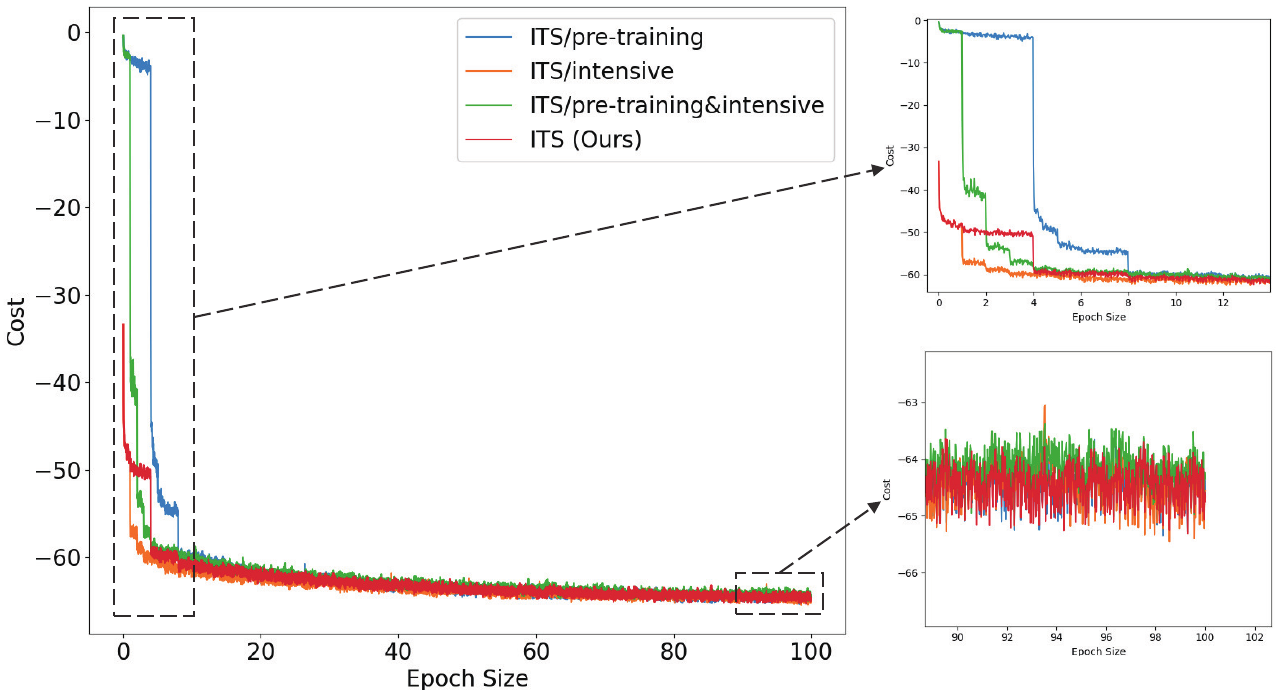}}
		\caption{Ablation study results of ITS}
		\label{Fig. 5}
	\end{figure}

	To investigate the efficacy of different processes in ITS, we construct the ablation study on the scenario U4 with 80 tasks. Different combinations of the training strategies are used, and we record the cost value during the training process in Fig. \ref{Fig. 5}. Particularly, “ITS/X” refers to our ITS but without the “X” component. For example, “ITS/intensive” denotes the training strategy that only includes the pre-training and alternate training processes.
	
	From Fig. \ref{Fig. 5}, we can observe that the pre-trained models perform better at the beginning of the interactive training (e.g., “ITS” versus “ITS/pre-training”). The intensive training process slows down the convergence by several epochs (e.g., “ITS/pre-training” versus “ITS/pre-training\&intensive”). However, as indicated in Table \ref{Table 1}, for each epoch, the time used to generate the training data for the lower-level model is much longer  than that of the lower-level model training (e.g., 116 minutes versus 31 minutes for U6-500). The intensive training process does not need this time-consuming training data generation for the lower-level model, which considerably reduces the whole training time. With respect to the eventual performance, our ITS exhibits a similar result to ITS/intensive, whereas the whole training time of ITS (ours) is much shorter than that of ITS/intensive due to the less lower-level training data generation. Thus, we use the proposed ITS which comes with an intensive training process to accelerate the entire training process without performance drop.

	\subsection{Generalization to Larger-scale Instances}
	
	To verify the generalization performance of our DL-DRL, we use the trained models to solve larger-scale instances. The DL-DRL (Sampling) is adopted since the sampling strategy can obtain better solutions with only slightly longer time than the greedy one. We first generalize the trained DL-DRL models from subsection V-B to solve the problem with a larger task scale. The results are shown in Fig. 6 and 7, respectively, where the horizontal coordinate refers to the problems that need to be solved, and the legend denotes the trained models. For example, N-100 represents the problem with 100 tasks, and U4-80 represents the DL-DRL model trained on U4 with 80 tasks.
	
\begin{figure}[htb]
		\centering{\includegraphics[width=3.4in]{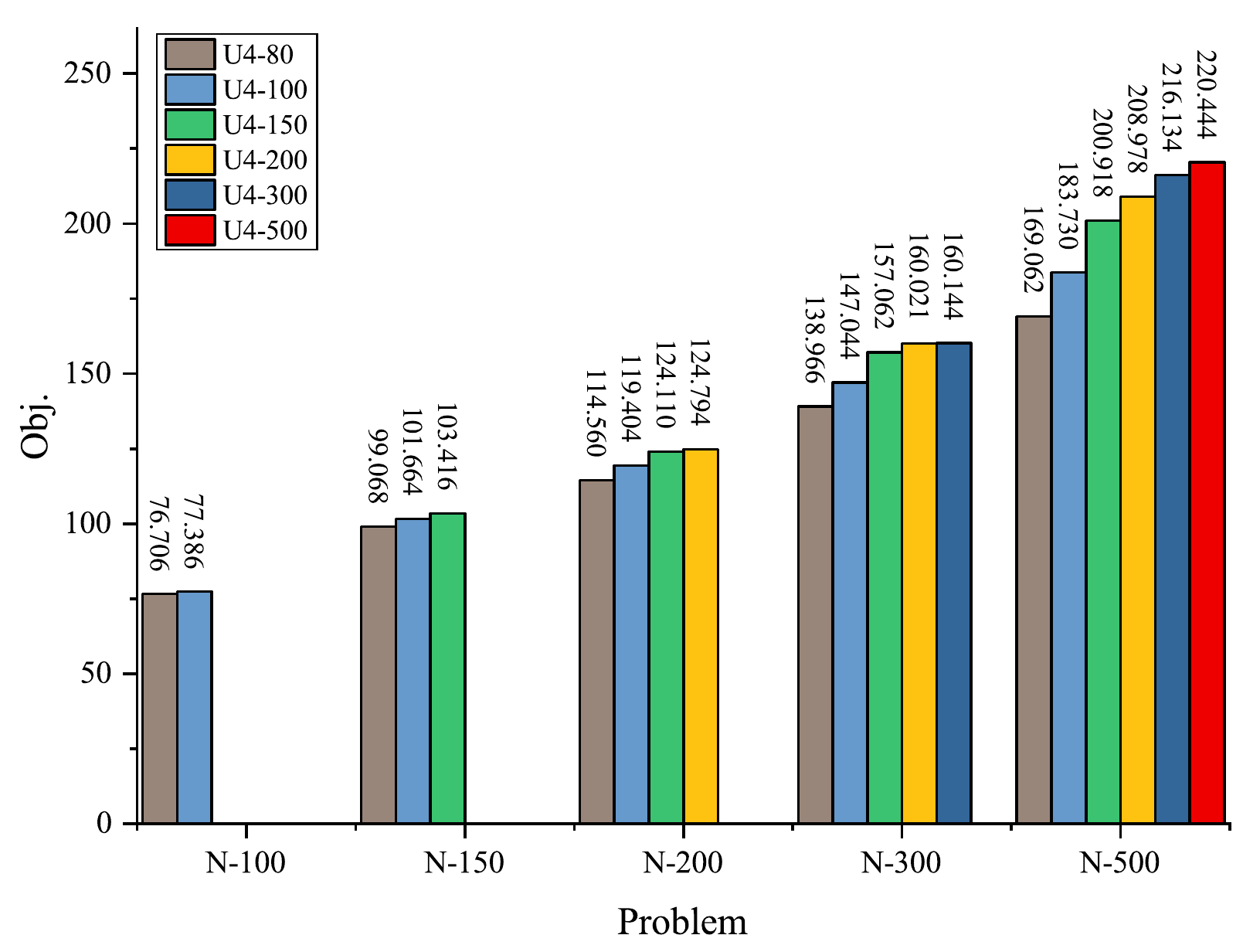}}
		\caption{Generalization to larger instances on U4}
		\label{Fig. 6}
		\vspace{-1.5pc}
	\end{figure}

\begin{figure}[htb]
		\centering{\includegraphics[width=3.4in]{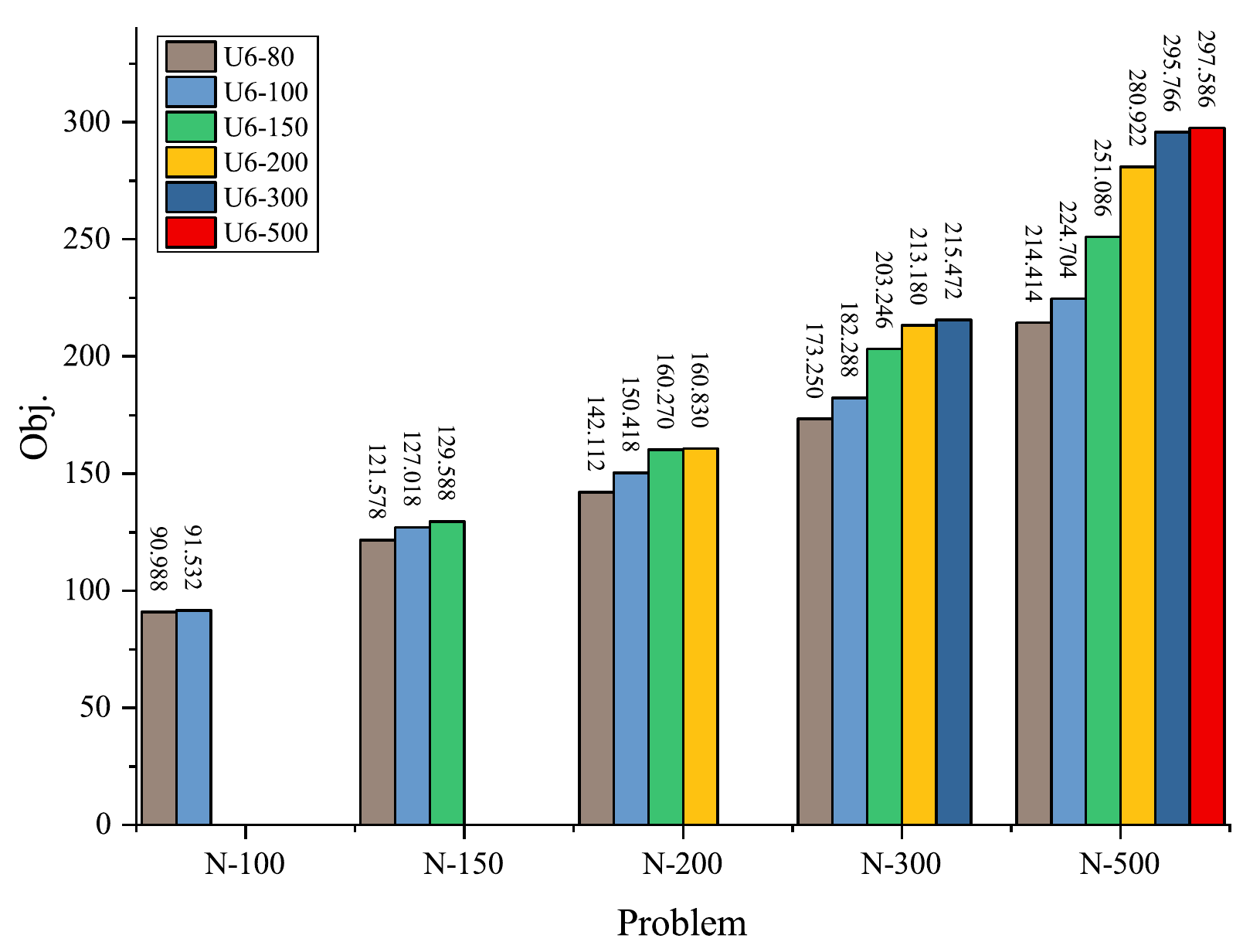}}
		\caption{Generalization to larger instances on U6}
		\label{Fig. 7}
	\end{figure}

	From Fig. \ref{Fig. 6}, we can see that the model trained for a certain problem size performs best on the corresponding problem compared to those trained for other problem sizes. However, they still outperform K-means+VND, K-means+AM, and DL-DRL-I except for U4-80 in solving problems with no less than 200 tasks and U4-100 in solving the problem with 500 tasks. In addition, we notice that the model trained for proximal problem sizes performs better than that of different ones (e.g., U4-200 and U4-300 perform better than U4-80, U4-100, and U4-150 in solving problems with 500 tasks). This phenomenon might be caused by the difference in data distribution, as the proximal problem sizes may lead to similar data distributions of task location. A similar pattern can also be found in Fig. \ref{Fig. 7}, where the models trained for other problem sizes outperform K-means+VND and K-means+AM and are competitive against DL-DRL-I.
	
	For larger-scale multi-UAV task scheduling problems with like 1000 tasks, it is intractable to train the model from scratch, rendering the generalizability a highly desired capability for the trained models. To further investigate the generalization performance of the proposed DL-DRL, we apply the one trained with 500 tasks to solve the instances with 600, 700, 800, 900, and 1000 tasks, and also compare it with the baselines including OR-Tools, K-means+AM, and DL-DRL-I. Table \ref{Table 4} displays the testing results of our DL-DRL and the baselines, including Obj. and average computation time. Compared to K-means+AM and DL-DRL-I, a salient superiority of DL-DRL can be observed in terms of Obj. with slightly longer computation time. Regarding OR-Tools, its computation time increases rapidly as the task scales up, and our DL-DRL exceeds it in terms of both Obj. and computation time. On all the scenarios, our DL-DRL achieves the best overall performance among all studied methods, which justifies its favorable generalization capability.
	
	\begin{table*}[htbp]
		\renewcommand\arraystretch{1.2}
		\footnotesize
		\centering
		\caption{Generalization results of DL-DRL and baselines}
		\begin{tabular}{c|c|cc|cc|cc|cc|cc}
			\toprule
			\multicolumn{1}{c|}{\multirow{2}[2]{*}{}} & \multirow{2}[1]{*}{Method} & \multicolumn{2}{c|}{N=600} & \multicolumn{2}{c|}{N=700} & \multicolumn{2}{c|}{N=800} & \multicolumn{2}{c|}{N=900} & \multicolumn{2}{c}{N=1000} \\
			\multicolumn{1}{c|}{} & \multicolumn{1}{c|}{} & \multicolumn{1}{c}{Obj.} & \multicolumn{1}{c|}{Time(s)} & \multicolumn{1}{c}{Obj.} & \multicolumn{1}{c|}{Time(s)} & \multicolumn{1}{c}{Obj.} & \multicolumn{1}{c|}{Time(s)} & \multicolumn{1}{c}{Obj.} & \multicolumn{1}{c|}{Time(s)} & \multicolumn{1}{c}{Obj.} & \multicolumn{1}{c}{Time(s)} \\
			\midrule
			\multicolumn{1}{c|}{\multirow{4}[1]{*}{U4}} & OR-tools & 231.79 & 16.93 & 251.65 & 22.15 & 271.20 & 28.50 & 288.36 & 36.17 & 305.64 & 43.01 \\
			& K-means+AM & 165.73 & \textbf{0.27} & 174.75 & \textbf{0.28} & 181.54 & \textbf{0.29} & 187.72 & \textbf{0.30}  & 192.43 & \textbf{0.31} \\
			& DL-DRL-I & 206.46 & 1.06  & 220.48 & 1.31  & 232.94 & 1.61  & 241.13 & 1.93  & 248.81 & 2.28 \\
			& DL-DRL & \textbf{245.67} & 1.07  & \textbf{267.88} & 1.33  & \textbf{287.36} & 1.62  & \textbf{303.83} & 1.95  & \textbf{321.58} & 2.34 \\
			\midrule
			\multicolumn{1}{c|}{\multirow{4}[1]{*}{U6}} & OR-tools & 326.37 & 30.45 & 356.94 & 39.70 & 386.45 & 52.30 & 411.60 & 65.73 & 437.17 & 79.83 \\
			& K-means+AM & 209.12 & \textbf{0.35} & 221.79 & \textbf{0.37} & 230.78 & \textbf{0.38} & 238.52 & 0.39  & 244.79 & \textbf{0.39} \\
			& DL-DRL-I & 254.55 & 1.47  & 277.81 & 1.78  & 295.52 & 2.21  & 305.69 & 2.67  & 316.94 & 3.24 \\
			& DL-DRL & \textbf{334.28} & 1.43  & \textbf{367.61} & 1.84  & \textbf{394.90} & 2.28  & \textbf{419.26} & 2.74  & \textbf{443.20} & 3.32 \\
			\bottomrule
		\end{tabular}%
		\label{Table 4}%
	\end{table*}%

	\section{Conclusion}
	
	In this paper, we investigate the DRL approach for the large-scale multi-UAV task scheduling problem. Based on the divide and conquer framework (DCF), the original problem is decomposed into task allocation subproblem and UAV route planning subproblem, and we proposed a double-level DRL (DL-DRL) approach to solve them. In our DL-DRL, two different DRL models are exploited to solve the two subproblems in the upper level and the lower level, respectively. Considering the intrinsic relationship of subproblems, an interactive training strategy (ITS) is designed for effective training, which comprises pre-training, intensive training, and alternate training. The experimental results show that our DL-DRL achieves the best overall performance compared to exact solver, conventional heuristic, and learning-based baselines, especially on large-scale instances. Furthermore, the generalization of our DL-DRL to larger-scale problems and the effectiveness of our ITS are also verified through experiments, respectively. Regarding the future studies, we will further improve our approach to tackle various distributions of task locations, test on real-world instances, and compare with other strong heuristic methods.

	\section{Acknowledgments}
	
	This work was supported in part by the National Natural Science Foundation of China under Grants 62073341 and the Fundadmental Reasearch Funds for the Central Universities of Central South University (No. 2022ZZTS0659).


	\bibliographystyle{IEEEtran}
	\bibliography{IEEEabrv, mybib}
	
	\begin{IEEEbiography}[{\includegraphics[width=1in,height=1.25in,clip,keepaspectratio]{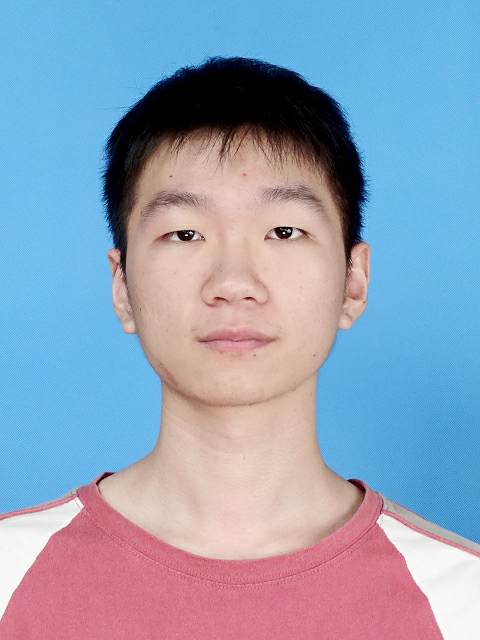}}]{Xiao Mao}{\space}
		received the B.S. degree in transportation engineering from Central South University, Changsha, China, in 2020. He is currently pursuing the Ph.D. degree in School of Traffic and Transportation Engineering, Central South University. His research interests include reinforcement learning and intelligent scheduling.
	\end{IEEEbiography}
	
	\begin{IEEEbiography}[{\includegraphics[width=1in,height=1.25in,clip,keepaspectratio]{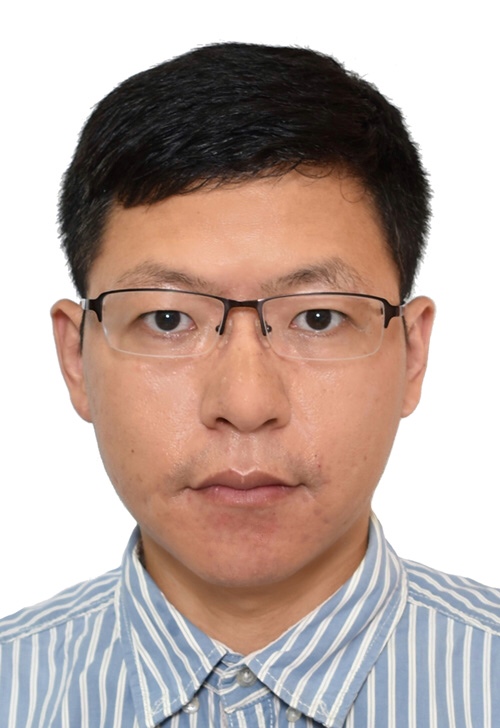}}]{Zhiguang Cao} received the Ph.D. degree from Interdisciplinary Graduate School, Nanyang Technological University. He received the B.Eng. degree in Automation from Guangdong University of Technology, Guangzhou, China, and the M.Sc. in Signal Processing from Nanyang Technological University, Singapore, respectively. He was a Research Fellow with the Energy Research Institute @ NTU (ERI@N), a Research Assistant Professor with the Department of Industrial Systems Engineering and Management, National University of Singapore, and a Scientist with the Agency for Science Technology and Research (A*STAR), Singapore. He joins the School of Computing and Information Systems, Singapore Management University, as an Assistant Professor. His research interests focus on learning to optimize (L2Opt).
	\end{IEEEbiography}
	
	\begin{IEEEbiography}[{\includegraphics[width=1in,height=1.25in,clip,keepaspectratio]{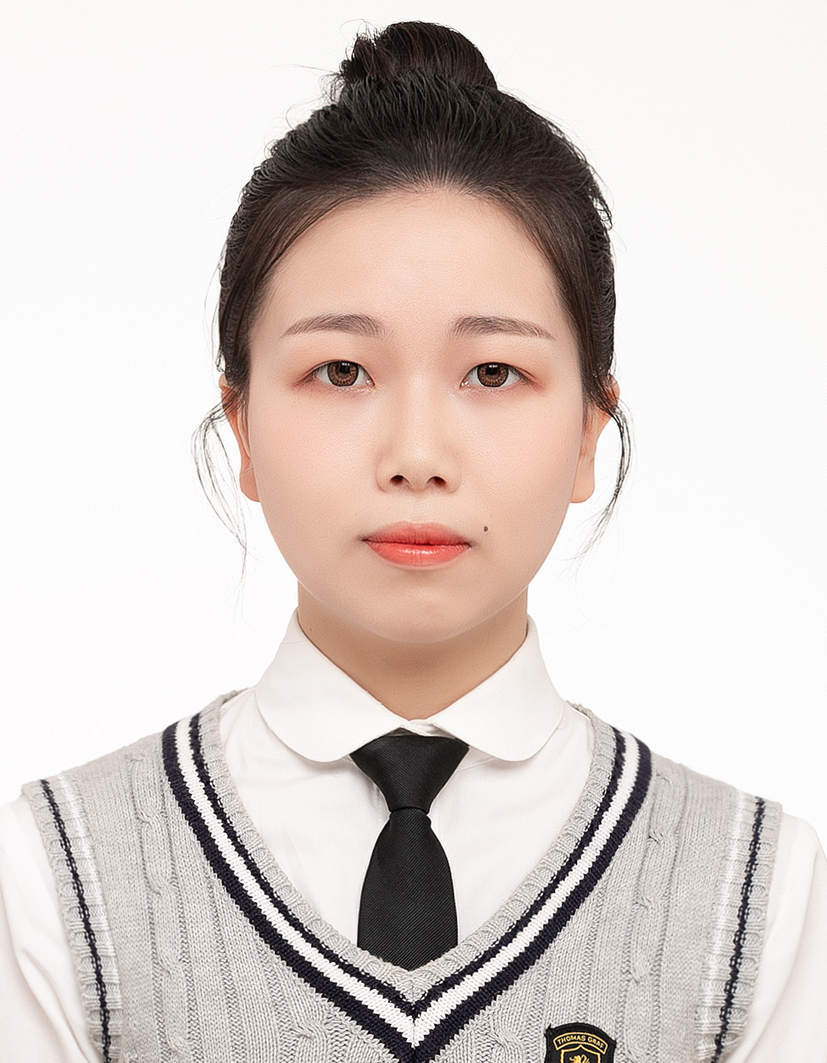}}]{Mingfeng Fan}{\space}
		received the B.S. degree in transport equipment and control engineering from Central South University, Changsha, China, in 2019, where she is currently pursuing the Ph.D. degree in traffic and transportation engineering. Her research interests include machine learning and UAV path planning.
	\end{IEEEbiography}
	
	\begin{IEEEbiography}[{\includegraphics[width=1in,height=1.25in,clip,keepaspectratio]{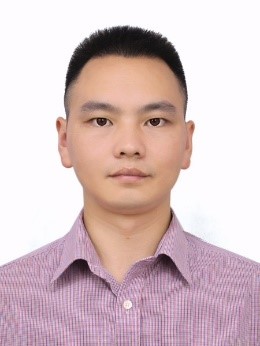}}]{Guohua Wu} (Member, IEEE) {\space}
		received the B.S. degree in Information Systems and Ph.D degree in Operations Research from National University of Defense Technology, China, in 2008 and 2014, respectively. During 2012 and 2014, he was a visiting Ph.D student at University of Alberta, Edmonton, Canada. He is now a Professor at the School of Traffic and Transportation Engineering, Central South University, Changsha, China. 
		
		His current research interests include scheduling, computational intelligence, and machine learning. He has authored more than 100 referred papers including those published in \textit{IEEE TCYB}, \textit{IEEE TEVC} and \textit{IEEE TSMCA}. He serves as an Associate Editor of \textit{Information Sciences}, and \textit{Swarm and Evolutionary Computation}, an editorial board member of \textit{International Journal of Bio-Inspired Computation}, a Guest Editor of \textit{Information Sciences and Memetic Computing}. 
	\end{IEEEbiography}

	\begin{IEEEbiography}[{\includegraphics[width=1in,height=1.25in,clip,keepaspectratio]{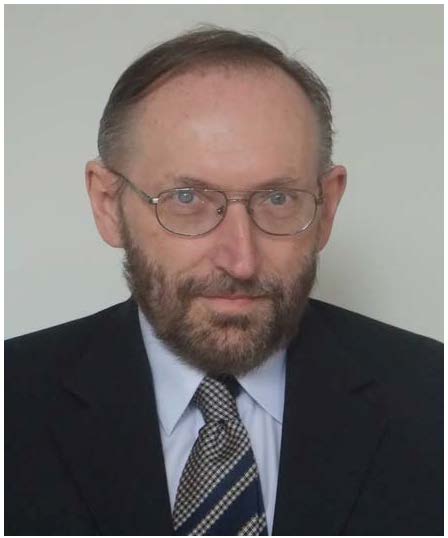}}]{Witold Pedrycz} (Life Fellow, IEEE) {\space}
		is a Professor and the Canada Research Chair (CRC-Computational Intelligence) with the Department of Electrical and Computer Engineering, University of Alberta, Edmonton, AB, Canada, and also with the Department of Electrical and Computer Engineering, Faculty of Engineering, King Abdulaziz University, Jeddah, Saudi Arabia. He is also with the Systems Research Institute, Polish Academy of Sciences, Warsaw, Poland. In 2012 he was elected a Fellow of the Royal Society of Canada. 
		
		His current research interests include computational intelligence, knowledge discovery and data mining, pattern recognition, knowledge-based neural networks, and software engineering. He has published numerous papers in the above areas. He is an Editor-in-Chief of \textit{Information Sciences}. He currently serves as an Associate Editor of \textit{IEEE Transactions on Fuzzy Systems} and \textit{IEEE Transactions on Systems, Man and Cybernetics: System} and is a member of a number of editorial boards of other international journals.
	\end{IEEEbiography}
	
	\vfill
	
\end{document}